\documentclass[letterpaper]{article} 
\usepackage{aaai2026}  
\usepackage{times}  
\usepackage{helvet}  
\usepackage{courier}  
\usepackage[hyphens]{url}  
\usepackage{graphicx} 
\usepackage{natbib}  
\usepackage{caption} 
\usepackage{algorithm}
\usepackage{algorithmic}

\usepackage{newfloat}
\usepackage{listings}
\DeclareCaptionStyle{ruled}{labelfont=normalfont,labelsep=colon,strut=off} 
\floatstyle{ruled}
\newfloat{listing}{tb}{lst}{}
\floatname{listing}{Listing}
\usepackage{booktabs}
\usepackage{bbding}
\usepackage{multirow}
\usepackage{subfigure}
\usepackage{makecell}
\usepackage{longtable}
\usepackage{tcolorbox}

\usepackage{colortbl}
\definecolor{mycolor}{rgb}{0.925, 0.945, 0.980}
\definecolor{mygray}{rgb}{0.886, 0.886, 0.886}
\definecolor{mypink}{rgb}{0.992, 0.921, 0.929}

\title{ProBench: Benchmarking GUI Agents with Accurate Process Information}
\author {
    Leyang Yang\textsuperscript{\rm 1,2},
    Ziwei Wang\textsuperscript{\rm 1,2},
    Xiaoxuan Tang\textsuperscript{\rm 3},
    Sheng Zhou\textsuperscript{\rm 1}\thanks{Corresponding Authors.},\\
    Dajun Chen\textsuperscript{\rm 3},
    Wei Jiang\textsuperscript{\rm 3},
    Yong Li\textsuperscript{\rm 3}\footnotemark[1],
}
\affiliations {
    \textsuperscript{\rm 1}Zhejiang Key Laboratory of Accessible Perception and Intelligent Systems, Zhejiang University\\
    \textsuperscript{\rm 2}College of Computer Science and Technology, Zhejiang University\\
    \textsuperscript{\rm 3}Ant Group\\
    \{yangleyang, wangziwei98, zhousheng\_zju\}@zju.edu.cn\\
    \{tangxiaoxuan.txx, chendajun.cdj, liyong.liy, jonny.jw\}@antgroup.com
}

\newcommand{\model}{ProBench}

\begin{document}

\maketitle

\begin{abstract}
With the deep integration of artificial intelligence and interactive technology, Graphical User Interface (GUI) Agent, as the carrier connecting goal-oriented natural language and real-world devices, has received widespread attention from the community.
Contemporary benchmarks aim to evaluate the comprehensive capabilities of GUI agents in GUI operation tasks, generally determining task completion solely by inspecting the final screen state. However, GUI operation tasks consist of multiple chained steps while not all critical information is presented in the final few pages. Although a few research has begun to incorporate intermediate steps into evaluation, accurately and automatically capturing this process information still remains an open challenge.
To address this weakness, we introduce \textbf{\model}, a comprehensive mobile benchmark with over 200 challenging GUI tasks covering widely-used scenarios. 
Remaining the traditional State-related Task evaluation, we extend our dataset to include Process-related Task and design a specialized evaluation method. A newly introduced Process Provider automatically supplies accurate process information, enabling presice assessment of agent's performance.
Our evaluation of advanced GUI agents reveals significant limitations for real-world GUI scenarios. These shortcomings are prevalent across diverse models, including both large-scale generalist models and smaller, GUI-specific models. A detailed error analysis further exposes several universal problems, outlining concrete directions for future improvements.
\end{abstract}


\section{Introduction}
With the deep integration of artificial intelligence~\cite{achiam2023gpt, gpt4v, Qwen2VL} and interactive technology, GUI Agent, as the carrier connecting natural language and real-world devices, has received widespread attention from the community~\cite{yao2024minicpm, hong2024cogagent, bai2024digirl, you2025ferret, wang2025mp}.
Early studies~\cite{refexp, widgetcaption, screenQA} concentrated on single-image problems, such as grounding and single-screen reasoning. As planning, reasoning and memory capabilities have advanced, recent works~\cite{wu2024atlas, lin2025showui} have explored more complex operation tasks with multiple steps.

\begin{figure}
    \centering
    \includegraphics[width=\linewidth]{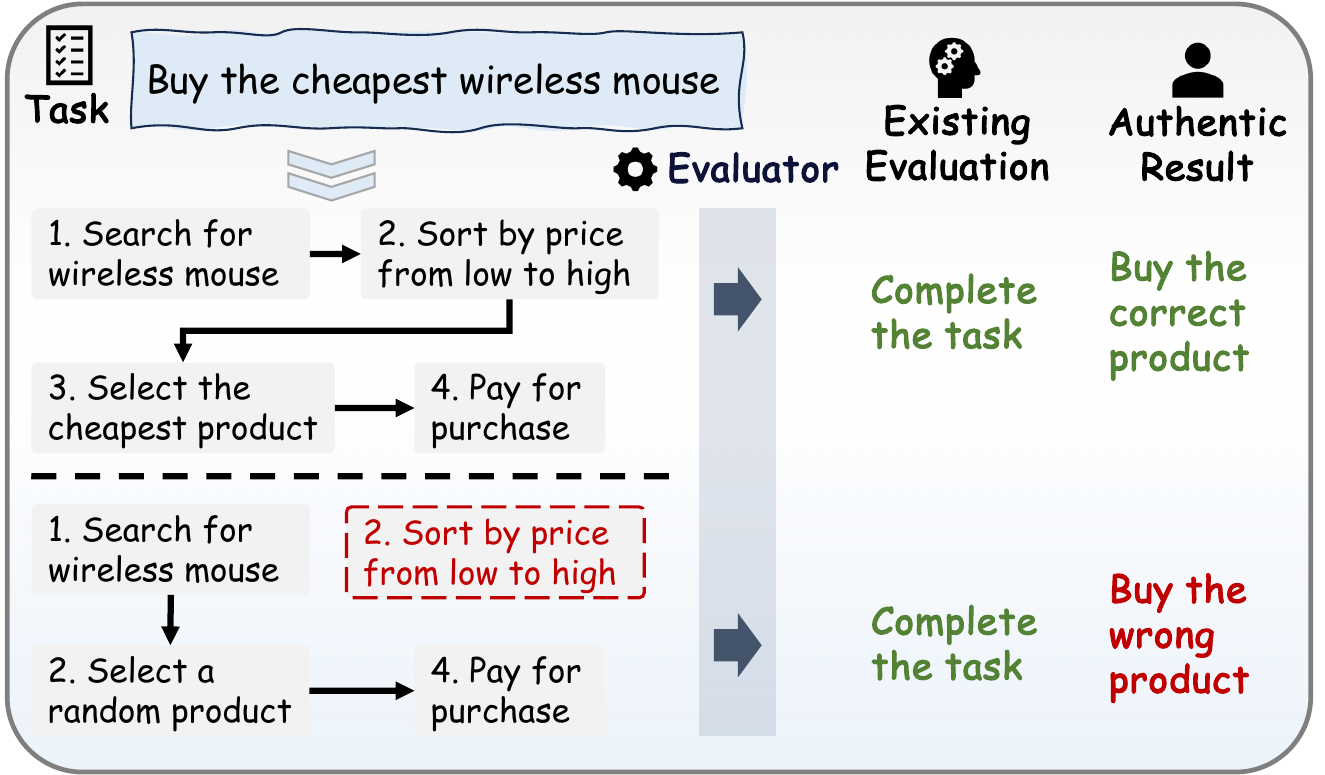}
    \caption{Illustration of a false outcome under existing evaluation. Red border indicates the ignored critical action.}
    \label{fig:op_task}
\end{figure}



However, by overlooking the inherently multi-step character of operation tasks, current GUI agents are typically evaluated only on the final screen state, leading to sub-optimal performance in practice.
Specifically, existing evaluations~\cite{lee2024benchmarkingmobiledevicecontrol, rawles2024androidworld, xu2024androidlab} generally determine the task completion exclusively relying on the final screen state, while not all critical information is contained in the final state.
Consider the illustrative task \textit{"Buy the cheapest wireless mouse"} in Figure~\ref{fig:op_task}, a crucial step is whether the agent explicitly complete \textit{"Sort by price from low to high"} before selecting. If the sorting step is omitted, the subsequent selection is effectively random, unlikely to choose the lowest-priced product. Under existing evaluation, both the properly sorted trajectory and the unsorted, erroneous trajectory are judged equally successful, because of the final screen showing a wireless mouse in the purchase confirmation view in either case. 



Recently, a few research has begun to consider intermediate process information into assessments. 
SPA-Bench~\cite{chen2024spa} downplays the importance of process information by explicitly decomposing high-level instructions into a rigid sequence of steps and guiding the agent to follow the pre-defined trajectory. Such a design neglects the evaluation of task planning capabilities. 
Furthermore, manually annotating inspectable intermediate states limits the task's scalability.
A3~\cite{chai2025a3} calls an LLM to decompose tasks into several essential states and judges the completion of subtasks through step-level comparisons, declaring the task complete only when all subtasks are fulfilled. 
However, the accuracy of this approach is constrained by the inherent limitations of existing LLM task decomposition.
Thus, accurately and automatically providing process information remains an open challenge.

To address this weakness, we introduce \textbf{\model}, a comprehensive mobile benchmark with over 200 challenging GUI tasks. \model\ covers widely-used scenarios of 34 mainstream Chinese and English online applications across media, news, social, shopping, etc.
For comprehensive evaluation, we retain the essential State-related Task, where all necessary information appears on the final screenshot. Therefore, performance can still be evaluated solely from the final screen state.
More importantly, to better reflect real-world scenarios, we include Process-related Task that place higher demand on the operation process. These tasks cannot be judged by the final screen alone. Accurate evaluation must consider both the final state and the critical operation steps (e.g., applying a price filter or specifying a delivery address).
To supply this information automatically, we introduce a Process Provider. Including Structure Description Converter which parses page hierarchy information while MLLM-based Summarizer employs MLLM to detect and summarize the changes by comparing screens, the two optional components provide process information accurately. With the above design, \model\ enables to evaluate the ability of GUI agents to rigorously capture and execute the necessary operation process. 

%

Our evaluation of advanced GUI agents on \model\ reveals significant limitations for real-world GUI scenarios. These shortcomings are prevalent across diverse models, including both large-scale generalist models and smaller, GUI-specific models. Besides, we figure out that GUI agents struggle with applications of social and lifestyle categories. An in-depth error analysis further uncovers several universal problems, offering directions for future improvements.
Our contributions are summarized as follows:
\begin{itemize}
    \item We introduce \model, a comprehensive mobile benchmark with over 200 challenging GUI tasks, covering widely-used scenarios across bilingual online apps.
    \item We design an automated evaluation pipeline, which provides critical process information through Process Provider, achieving efficient and accurate evaluation.
    \item The evaluation on \model\ reveals significant limitations in real-world scenarios, both large-scale generalist models and smaller, GUI-specific models. We also provide further analysis to uncover universal problems.
\end{itemize}

\begin{table}[]
\centering
\begin{tabular}{lcccc}
\toprule
Benchmark      & Online & Chinese & \makecell[c]{MV \\ Free} & \makecell[c]{AP \\ Process} \\ \midrule
AndroidArena & \XSolidBrush            & \XSolidBrush            & \XSolidBrush            & \XSolidBrush     \\
AndroidWorld & \Checkmark              & \XSolidBrush           & \XSolidBrush                 & \XSolidBrush\\
B-MoCA       & \Checkmark              & \XSolidBrush                         & \XSolidBrush                             & \XSolidBrush            \\
AndroidLab   & \XSolidBrush                        & \XSolidBrush                         & \XSolidBrush                             & \XSolidBrush            \\
SPA-BENCH    & \Checkmark                        & \Checkmark                        & \Checkmark                             & \XSolidBrush            \\
A3           & \XSolidBrush                        & \XSolidBrush                        & \Checkmark                             & \XSolidBrush            \\ \midrule
\textbf{\model}       & \Checkmark                        & \Checkmark                        & \Checkmark                             & \Checkmark            \\ \bottomrule
\end{tabular}
\caption{\textbf{Comparison of \model\ with other dynamic GUI benchmarks.} Online means whether includes third-party online apps and Chinese means whether includes Chinese apps. MV Free means whether the evaluation phase is free of manual verification participation. AP Process means whether capture precise process information automatically.}
\label{table:comparison}
\end{table}

\section{Related Work}
\subsection{Static GUI Benchmark}
RICO~\cite{deka2017rico} marked an important milestone in GUI-related research by providing a basic dataset for GUI element classification and detection. 
Afterwards, UGIF~\cite{venkatesh2022ugif} introduced instruction-based GUI control task. 
AITW~\cite{rawles2023androidinthewild} expanded the field with a large-scale dataset, while suffering from instruction redundancy and frequent mislabeling. AITZ~\cite{zhang2024android} refined AITW by applying Chain-of-Action-Thought reannotation and got a cleaner but smaller dataset. 
AndroidControl~\cite{li2024effects} further introduced a dataset with simpler tasks and unique action space compared to AITW. 
However, these static benchmarks evaluate agents by one-step operation based on single screenshot and correct action history. Without realistic historical actions, the method ignores the inherently dynamic and interactive feature of real-world environment. In addition, one-step error directs to task failure, leaving no opportunity to assess the capability of reflection or recovery.

\begin{figure*}
    \centering
    \includegraphics[width=\linewidth]{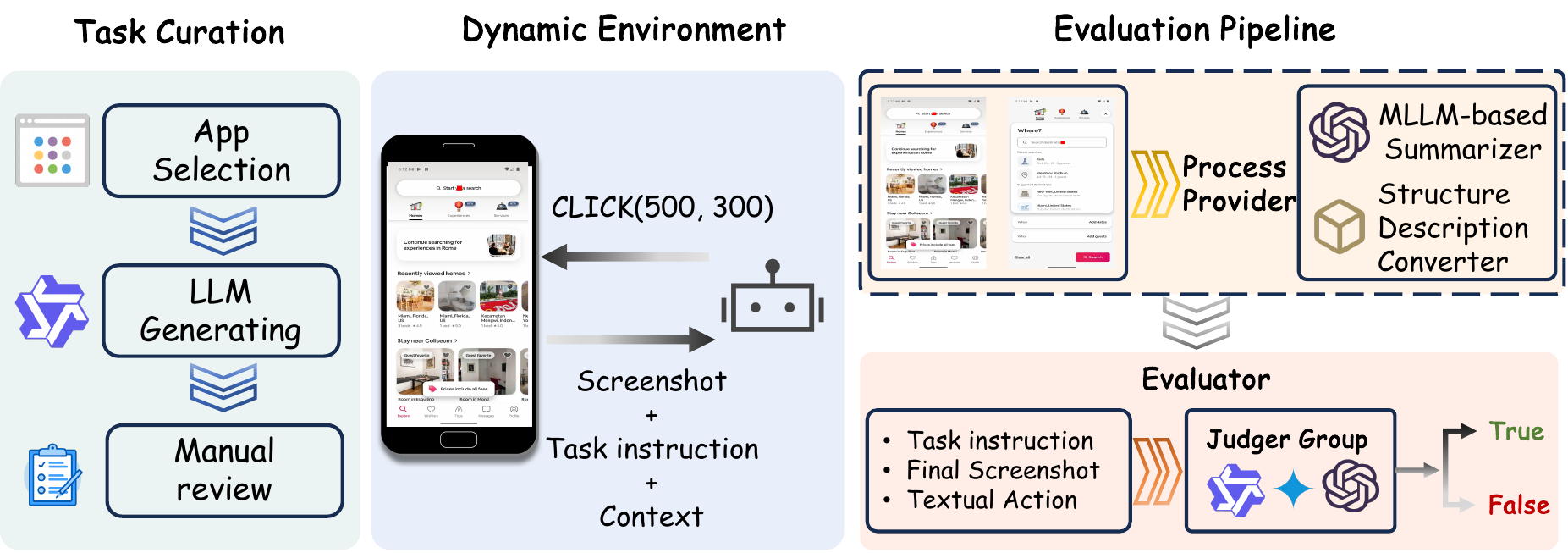}
    \caption{\textbf{Overview of our \model\ benchmark.} \model\ is a comprehensive mobile benchmark, comprising three key modules: \textbf{(\textit{i}) Task Curation:} We select 34 mainstream bilingual applications, generate candidate tasks using LLMs, and refine them through manual review. \textbf{(\textit{ii}) Dynamic Environment:} Agents complete the specified tasks by controlling the device. \textbf{(\textit{iii}) Evaluation Pipeline:} For Process-related Task, we optionally choose either the Structure Description Converter or the MLLM-based Summarizer of Process Provider to supply process information. The final evaluation is performed by the judger selected from the judger group.}
    \label{fig:overview}
\end{figure*}

\subsection{Dynamic GUI Benchmark}
Researchers have developed a range of dynamic evaluation systems to better approximate real-world environment. AndroidArena~\cite{xing2024understanding} focuses exclusively on Google and built-in system apps, limiting its capacity to evaluate the general performance of third-party apps. B-Moca~\cite{lee2024benchmarkingmobiledevicecontrol} incorporates applications such as Walmart, while the tasks are overly simplistic and lack diversity. AndroidWorld~\cite{rawles2024androidworld} increases task diversity by selecting open-source applications from F-Droid, while discrepancies in design formula remain between these applications and mainstream software. AndroidLab~\cite{xu2024androidlab} introduces additional variety by leveraging offline and static applications.
Concurrently, these benchmarks predominantly depend on element matching~\cite{lee2024benchmarkingmobiledevicecontrol} or predefined detection code~\cite{xu2024androidlab} on the final screen, omitting the evaluation of the crucial intermediate steps.
Realizing the shortcoming, SPA-BENCH~\cite{chen2024spa} and A3~\cite{chai2025a3} consider intermediate process information into assessments, while introducing MLLMs to assist the evaluation.
Nevertheless, these evaluations cannot capture process information accurately and automatically. \model\ addresses this gap by building a dedicated evaluation pipeline while capturing process information using Process Provider. Table~\ref{table:comparison} compares our work with existing dynamic GUI agent benchmarks, highlighting our innovations and unique contributions.
\section{\model}
\subsection{Overview}
Serving as the communication bridge between GUI agent and device, \model\ is implemented based on adbutils\footnote{https://github.com/openatx/adbutils}, an open-source python Android Debug Bridge(ADB) library, as illustrated in Figure~\ref{fig:overview}.

\model\ acquires the current screen state by capturing the real-time screenshot. The screenshot, together with the task instruction and context information (such as historical operation records), is transmitted to the agent. Upon receiving these inputs, the agent predicts the subsequent specific operation.
Textual operations are relayed back to \model, which parses and converts them into device control commands. These commands are executed via adbutils to manipulate the device. This iterative process continues until the agent confirms task completion or the system-defined maximum step limit is reached.

During the evaluation phase, evaluator receives the task instruction and final screenshot of the execution process. Concurrently, if the task is Process-related, users are afforded the flexibility to choose Structure Description Converter or MLLM-based Summarizer from Process Provider to provide accurate process information. The evaluator make final judgments under the input assistance.
\subsection{Task Curation}
\label{subsec: task}
Starting from the application list provided by SPA-BENCH~\cite{chen2024spa}, we remove (i) English applications that cannot be executed on virtual devices and (ii) applications that embed bot-detection or other anti-automation mechanisms.

The filtering process yields 34 target applications contained 14 English and 20 Chinese, covering a wide spectrum of categories for potential GUI scenarios.
For every application we first compose a handful of seed tasks manually. Leveraging the generative capacity of Qwen3~\cite{yang2025qwen3}, we then expand these seeds into a larger candidate pool. 
All automatically generated tasks undergo human filter and editing to guarantee correctness and diversity. In particular, we make efforts to  reduce lexical and structural overlap between instructions so that agents must exhibit genuine capabilities rather than memorize recurring patterns.
We define these operation tasks into two complementary types: State-related Task and Process-related Task, to ensure the diversity of our benchmark.

\textbf{State-related Task} focuses exclusively on the final screen state. The task is deemed successful if the requested information is visible on the screen in the end, regardless of the intermediate action sequence. For instance, in the task \textit{"Query the current balance in Alipay"}, the task is successful once the correct balance amount is shown on the screen. The specific sequence of actions or operations employed is immaterial if the final state meets the task instruction.

\begin{figure}
    \centering
    \includegraphics[width=\linewidth]{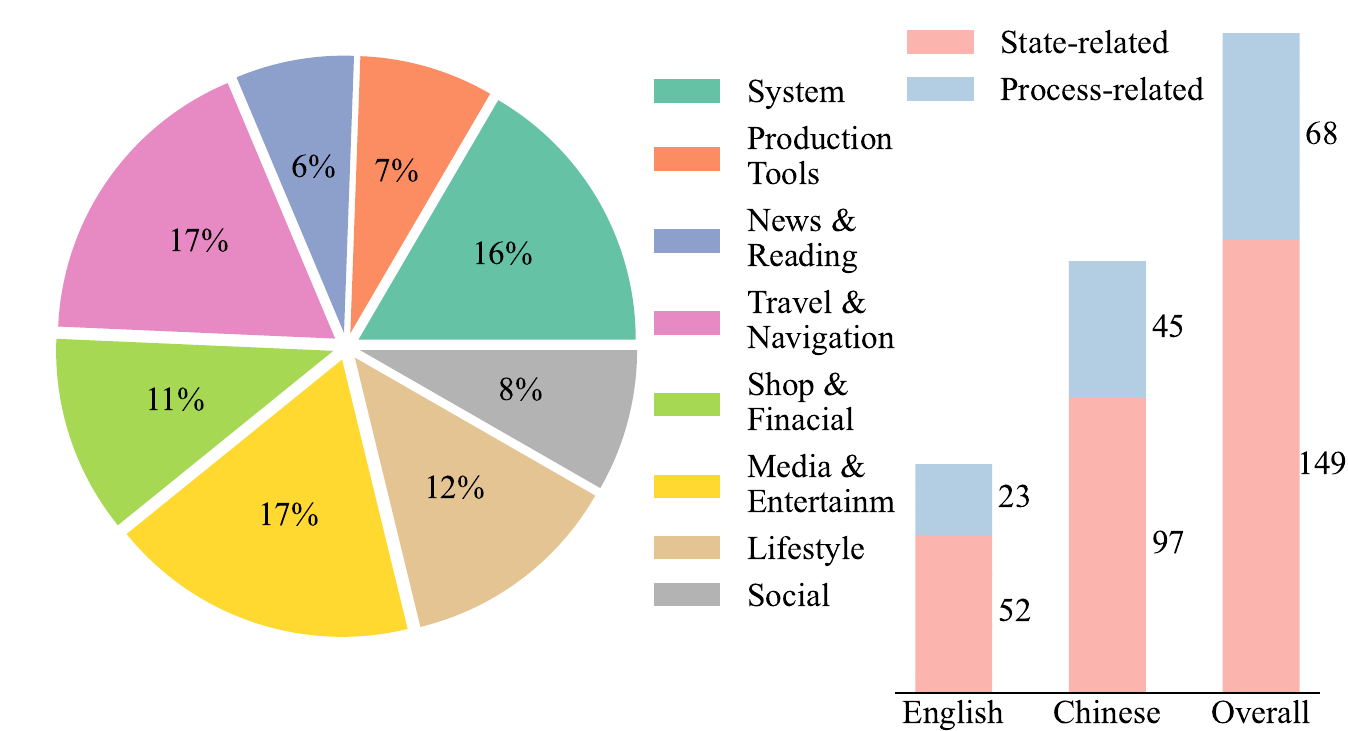}
    \caption{\textbf{Statistics of \model.} The left illustrates the diverse application categories included. The right displays the amount of each type tasks within different languages.}
    \label{fig:composition}
\end{figure}

\textbf{Process-related Task} demands a more comprehensive consideration: the agent must execute some specific operation process, not just reach the end state. Considering the instruction \textit{"Find the highest-rated sushi restaurant in Tokyo and check its full menu"}, merely displaying the menu is insufficient. The agent must also perform a series of implicit operations that are not directly observable from the final screen state, such as locating Tokyo, sorting restaurants by ratings, and selecting sushi restaurants. Both the final screen and the intervening critical steps are therefore evaluated.
Figure~\ref{fig:composition} shows the statistics of \model.


\subsection{Action Space}
In alignment with the action space defined in AITW, \model\ inherits the primitive actions, including {\tt CLICK, SWIPE, TYPE, ENTER, BACK, COMPLETE}. Given that all tasks are confined to single application, we remove the {\tt HOME} action from the action space, disallowing navigation to other applications. Considering the inherent challenges posed by network latency in online services, we additionally introduce a {\tt WAIT} action to handle situations where network resources are slow to load, allowing the system to pause execution.

\subsection{Evaluation Pipeline}
We define three possible final outcomes in our evaluation pipeline: (i) \underline{\textbf{Uncompleted}} (the agent reaches the step budget without emitting the completion signal), (ii) \underline{\textbf{Failure}} (the agent does emit completion, but the completed actions does not satisfy the task requirements), and (iii) \underline{\textbf{Success}}(the agent emits completion and achieves the task goal).

For State-related Task, we retain the evaluation method by relying solely on the final screen state. Once the agent issues the task completion, the evaluator determine whether the task instruction is fulfilled via the last screenshot.

As noted in Section~\ref{subsec: task}, Process-related Task relies on both the final state and the intermediate operations which is called process information. To provide precise process information, we develop the Process Provider composed with two optinal components:
\begin{enumerate}
\item \textbf{Structure Description Converter}\quad As shown in Algorithm~\ref{alg:xml}, after every click, we parse the a11y tree and locate the smallest clickable node enclosing the click coordinates. We record its \textit{text} and \textit{content\_desc} attributes. If both are empty, we extract the \textit{resource\_id}, which often indicates the functional hint such as \textit{filter\_icon}, then supplement it with details of relevant child nodes. The resulting string serves as a compact, human-readable description of the action.

\begin{algorithm}[tb]
\caption{Convert Structure Information to Description}
\label{alg:xml}
\textbf{Input}: a11y\_tree, click\_coordinate\\
\textbf{Output}: desc
\begin{algorithmic}[1] 
\STATE structure $\gets$ parse\_tree(a11y\_tree)
\STATE node $\gets$ get\_minimum(structure, click\_coordinate)
\STATE desc $\gets$ parse\_description(node)
\IF {desc is empty}
\STATE desc $\gets$ parse\_resourceid(node)
\FOR{child in node}
\STATE desc $\gets$ desc + parse\_description(child)
\ENDFOR
\ENDIF

\STATE \textbf{return} desc
\end{algorithmic}
\end{algorithm}

\item \textbf{MLLM-based Summarizer}\quad Because the structure information can be noisy or missing arising from software design variances, we include MLLM-based Summarizer, employing MLLM to compare the screenshots before and after the action to generate a descriptive textual summary. Specifically, we merge the two screenshots before and after the {\tt CLICK} action and mark the click coordinate in the image. MLLM-based Summarizer compare the screenshots and generates a descriptive textual summary of the performed operation, as shown in Figure~\ref{fig:summary}.

\begin{figure}[]
    \centering
    \includegraphics[width=\linewidth]{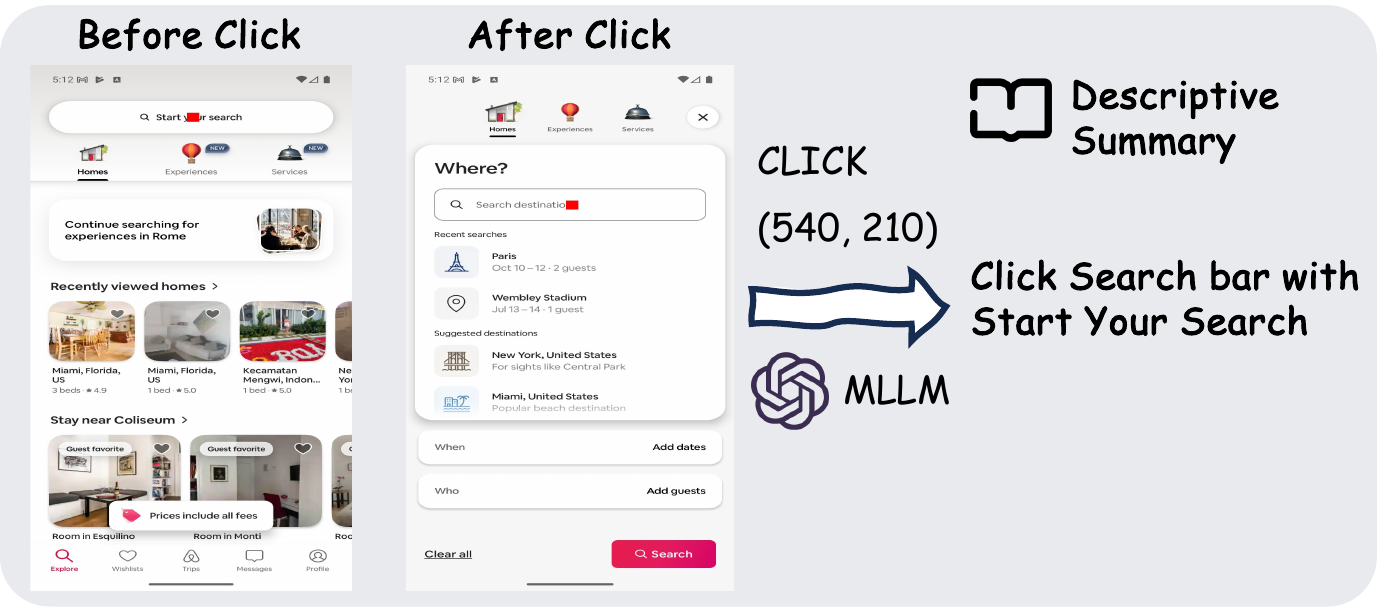}
    \caption{The implementation example of MLLM-based Summarizer. Take clicking the search box on the homepage of Airbnb as the example.}
    \label{fig:summary}
\end{figure}

\end{enumerate}
The evaluator receives the complete sequence of textualized actions together with the final screenshot to judge whether the key procedure has been performed and the task completed correctly.

A comparison with human annotations confirms the reliability of our evaluation method. Detailed accuracy are reported in Table ~\ref{table:efficiency}.

\begin{table}[]
\centering
\renewcommand{\arraystretch}{1.2}
\begin{tabular}{@{}llc@{}}
\toprule
Evaluation Method      & Task Type           & Correct \\ \midrule
Evaluator                   & State-related                  & 96.0\%  \\
\makecell[l]{Evaluator + \\ Structure Description Converter} & Process-related & 89.7\%  \\
\makecell[l]{Evaluator + \\ MLLM-based Summarizer} & Process-related                    & 94.1\%  \\ \bottomrule
\end{tabular}
\caption{\textbf{The correctness of our evaluation pipeline.} We choose Gemini 2.5 Pro~\cite{comanici2025gemini} as the MLLM for both the Evaluator and the Summarizer. "Correct" represents the correctness determined by human.}
\label{table:efficiency}
\end{table}
\begin{table*}[]
\centering
\renewcommand{\arraystretch}{1.2}
\setlength{\tabcolsep}{12pt}
\begin{tabular}{llllllllll}
\toprule
                        & \multicolumn{3}{c}{\textbf{English}}                            & \multicolumn{3}{c}{\textbf{Chinese}}                            & \multicolumn{3}{c}{\textbf{Overall}} \\ \cmidrule(l){2-10} 
\multirow{-2}{*}{\textbf{Model}} & \multicolumn{1}{c}{ST} & \multicolumn{1}{c}{PT} & Avg. & \multicolumn{1}{c}{ST} & \multicolumn{1}{c}{PT} & Avg. & ST      & PT     & Avg.     \\ \midrule
\multicolumn{10}{l}{\cellcolor[HTML]{EFEFEF}\textit{Proprietary Models}}                                                                                                 \\
GPT-4o                  & 0.0                       & 0.0                       & 0.0     & 0.0                       & 0.0                       & 0.0     & 0.0        & 0.0       & 0.0         \\
Claude 4 Sonnet                  & 0.0                       & 0.0                       & 0.0     & 0.0                       & 0.0                       & 0.0     & 0.0        & 0.0       & 0.0         \\
Gemini 2.5 Pro          & 44.2                       & 21.7                       & 37.3     & \cellcolor{mypink}\textbf{46.4}                       & \cellcolor{mypink}\textbf{31.1}                       & \cellcolor{mypink}\textbf{41.5}     & \cellcolor{mypink}\textbf{45.6}        & \cellcolor{mypink}\textbf{27.9}       & \cellcolor{mypink}\textbf{40.1}         \\ \midrule
\multicolumn{10}{l}{\cellcolor[HTML]{EFEFEF}\textit{General Open-source Models}} \\

Qwen2.5-VL-7B          & 7.7                       & 0.0                       & 5.3     & 6.2                       & 2.2                       & 4.9     & 6.7        & 1.5       & 5.1         \\
Qwen2.5-VL-32B          & 26.9                       & 8.7                       & 21.3     & 14.4                       & 13.3                       & 14.1     & 18.8        & 11.8       & 16.6         \\
Qwen2.5-VL-72B          & \cellcolor{mypink}{\textbf{63.5}}                       & \cellcolor{mypink}\textbf{30.4}                       & \cellcolor{mypink}\textbf{53.3}     & 28.9                       & 26.7                       & 28.2     & 40.9        & \cellcolor{mypink}\textbf{27.9}       & 36.9         \\
InternVL3-8B          & 13.5                       & 0.0                       & 9.3     & 3.1                       & 2.2                       & 2.8     & 6.7        & 1.5       & 5.1         \\
\midrule
\multicolumn{10}{l}{\cellcolor[HTML]{EFEFEF}\textit{GUI-specific Models}}                                                                                               \\
UI-TARS-1.5-7B              & 26.9                       & 8.7                       & 21.3     & 3.1                       & 0.0                       & 2.1     & 11.4        & 2.9       & 8.8         \\
UI-R1-E-3B              & 0.0                       & 0.0                       & 0.0     & 0.0                       & 0.0                       & 0.0     & 0.0        & 0.0       & 0.0         \\
GUI-R1-3B               & 15.4                       & 0.0                       & 10.7     & 10.3                       & 0.0                       & 7.0     & 12.1        & 0.0       & 8.3         \\ 
\bottomrule
\end{tabular}
\caption{Evaluation results of advanced GUI agents. The highest model performance in each column is highlighted in pink. ST is short for State-related Task and PT is short for Process-related Task.}
\label{table:main_result}
\end{table*}

\section{Experiments}
\subsection{Models}
In order to comprehensively evaluate the performance of current GUI agents, we select several representative agents covering three categories: proprietary models, general open-source models and GUI-specific models.
For proprietary models, we choose GPT-4o~\cite{hurst2024gpt}, Claude 4 Sonnet~\cite{drolet2023claude4} and
Gemini 2.5 Pro~\cite{comanici2025gemini}. For general open-source models, we choose Qwen2.5-VL~\cite{bai2025qwen2} with 7B, 32B, 72B and InternVL3-8B~\cite{zhu2025internvl3}. For GUI-specific models, we choose UI-TARS-1.5-7B~\cite{qin2025ui}, UI-R1-E-3B~\cite{lu2025ui} and GUI-R1-3B~\cite{luo2025gui}. The template of each model will be shown in Appendix. We do not incorporate Set-of-Mark~\cite{yang2023set}, so the model output coordinates directly.
\subsection{Evaluation Setup}
English applications are run on the Android emulator\footnote{https://developer.android.com/studio/run/emulator} and Chinese applications are executed on a physical Android phone equipped with AdbKeyboard\footnote{https://github.com/senzhk/ADBKeyBoard} to enable Chinese-character input.
Before every execution, we manually clear all in-app histories to guarantee a consistent initial state. Each task is allotted a maximum of 15 interaction steps.
For evaluation, we employ the Structure Description Converter to supply process information, and use Gemini 2.5 Pro as the judger in evaluator.

\subsection{Main Results}
\subsubsection{Overall performance}
Table~\ref{table:main_result} reports the evaluation results on our \model. Gemini 2.5 Pro ranks first with an average accuracy of 40.1\%, confirming its strong comprehensive capability.
Regardless of model size or training paradigm, performance is consistently higher on State-related Task than on Process-related Task, indicating that the latter which require dynamic attention to process, place higher demand on GUI agent. However, even for the simpler State-related Task, current agents still struggle: only Qwen2.5-VL-72B surpasses 60 \% accuracy on English State-related Task, highlighting the substantial room for improvement in GUI agents for real-world GUI scenarios.

\subsubsection{General open-source models exhibit a significant scaling effect.}
As the scale of model parameters increases, the Qwen2.5-VL series improves steadily, reaching an overall accuracy of 36.9\%, closing the gap with the level of top proprietary model. In English operation tasks, the 72B variant achieves an average accuracy of 53.3\%, outperforming every other model. These results underline that, in GUI scenarios, a larger model scale can yield substantial performance gains.

\subsubsection{GUI-specific models exhibit limited generalization.}
UI-R1-E-3B fails to complete even a single task, related to the lack of {\tt COMPLETE} action examples in its training phase. The agent never issues a completion signal even if it has completed the task correctly.
UI-TARS-1.5-7B, while markedly outperforming its base model Qwen2.5-VL-7B on English tasks, still lags far behind large-parameter general models with good grounding capability. These observations suggest that task-specific fine-tuning or reinforcement learning on static datasets has a natural gap with the real environment and cannot bridge deficiencies in basic vision language understanding. Although domain-specific data promotes performance improvement, robust and strong general foundational capabilities remain the decisive factor.

\subsection{Performance on Application Categories}
\begin{figure}
    \centering
    \includegraphics[width=\linewidth]{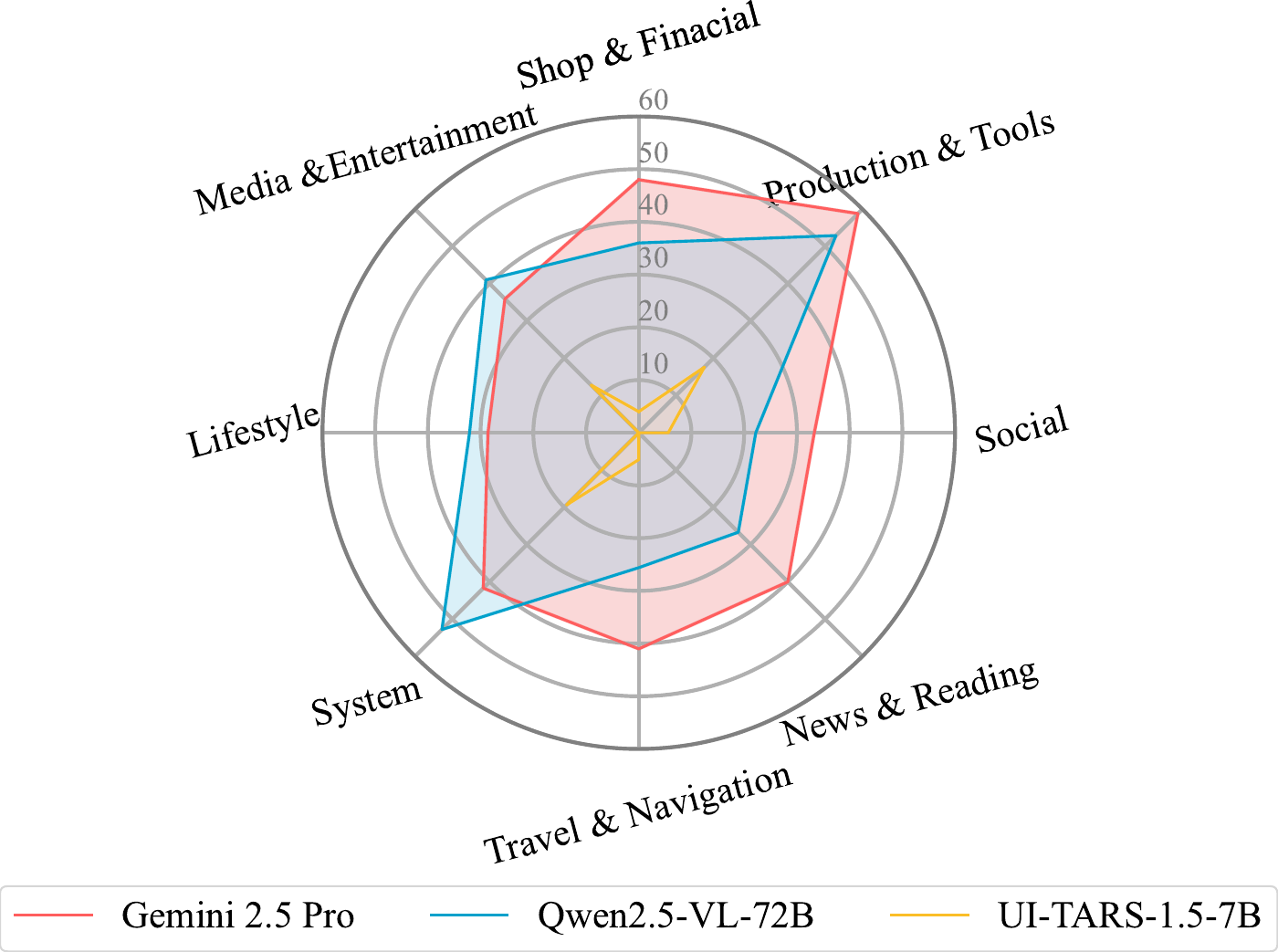}
    \caption{Accuracies of GUI agents on \model\ among different application categories. We demonstrate the best proprietary, general open-source and GUI-specific models.}
    \label{fig:accuracy}
\end{figure}

Figure~\ref{fig:accuracy} reports accuracy by application category. The agents generally perform better on production tools and system applications,  while struggling with social-networking and lifestyle applications, the very domains that most closely match daily information-retrieval requests. Typical user tasks include \textit{"show the latest direct message from user X"}, \textit{"how long until my food arrives"}, or \textit{"where is my package right now"}.
Several interface properties explain the gap. Social and lifestyle applications refresh content frequently and display information in a highly fragmented manner than relatively static pages like shopping or system settings: icon-only buttons, deeply nested folding cards, and advertising pop-ups crowd the limited screen estate. 
Efficient signals are sparse, while visual distractors abound, making it difficult for current GUI agents to complete instructions, ultimately depressing the overall accuracy.

\section{Error Analysis}
To illustrate the obstacles encountered by agents during execution, we present several error cases from our evaluation and distill them into universal problems.

\subsection{Insufficient Grounding Capability}
Grounding capability, the ability to accurately interpret and locate GUI elements, serves as the foundation for successful completion of various GUI tasks. However, most proprietary models lack pre-training on GUI-related data and consequently show obvious deficiencies in grounding, seriously constraining the potential for performance improvement.

\begin{figure}[]
  \centering
  \includegraphics[width=\linewidth]{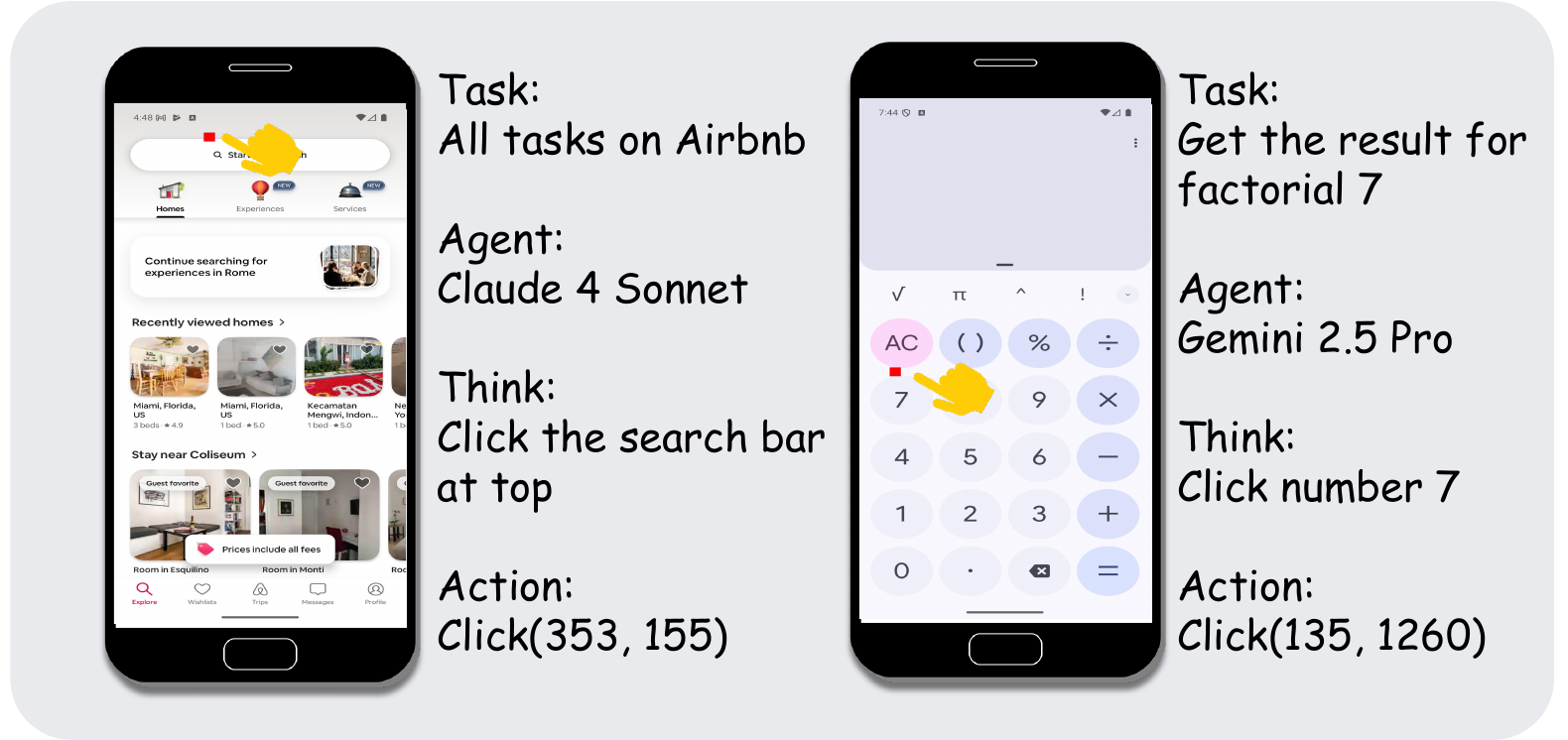}
  
  \caption{The lack of grounding capability in GUI agents. We show the accurate screen location where the agent actually clicked after coordinate conversion.}
  \label{fig:grounding}
\end{figure}

GPT-4o and Claude 4 Sonnet perform poorly in our evaluation, primarily due to their limited grounding capability. A critical barrier to task progression is their inability to accurately locate GUI elements, resulting in difficulties generating precise coordinates. As illustrated in Figure~\ref{fig:grounding}, Claude attempts to click the search box on the Airbnb homepage but fails to accurately locate it. Unable to complete the initial step ultimately leads to the failure of all tasks on Airbnb.
Gemini 2.5 Pro also experiences shortcomings in grounding capability. When assigned the task \textit{"Get the result for factorial 7"}, Gemini is unable to correctly select the number 7, directly causing the task to fail. These cases demonstrate that current mainstream proprietary models are generally insufficient in grounding ability, largely due to the absence of pre-training on specialized GUI-related knowledge.

\subsection{Insensitive to Historical Operation}
Compared to State-related Task, Process-related Task imposes higher demands on the tracking of historical operations, where the final screen state usually fails to fully represent all the requirements of the task. Agents not only need to comprehend the current screen state, but also integrate previous actions to accurately determine whether the task is truly completed. In other words, only by thoroughly considering the historical operations can agents make correct judgments regarding task completion.

\begin{figure}
    \centering
    \includegraphics[width=\linewidth]{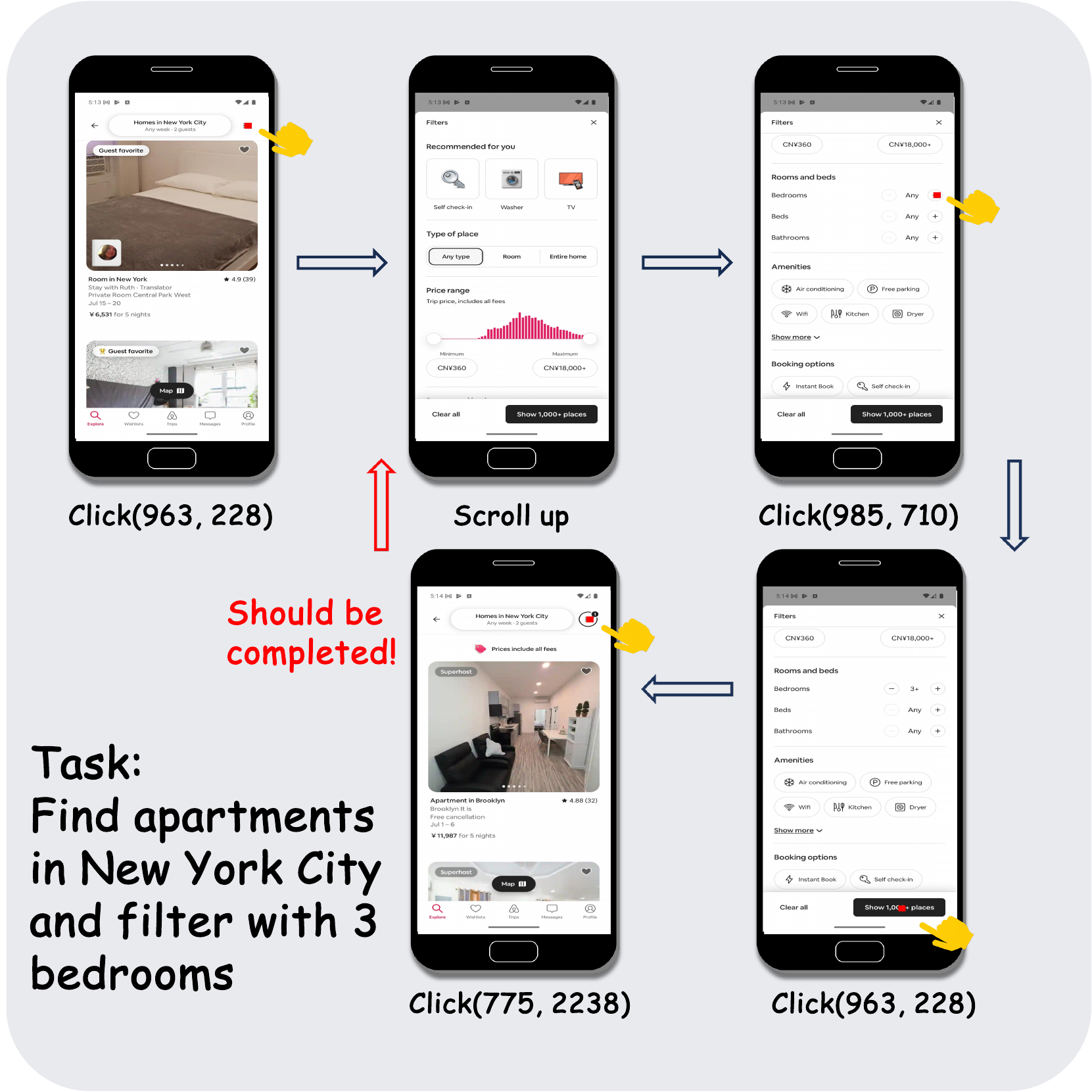}
    \caption{An example where GUI agent is insensitive to historical operation. Black arrows indicate correct action sequence, and red arrow indicates incorrect action that should complete.}
    \label{fig:history}
\end{figure}

However, most models fail to adequately attend to historical operations. As shown in Figure~\ref{fig:history}, we use the example of Qwen2.5-VL-72B performing the task \textit{"Find apartments in New York City and filter with 3 bedrooms"}, focusing particularly on its attempts to complete \textit{"filter with 3 bedrooms"}. In practice, due to Qwen's inability to effectively utilize historical operation information, it does not recognize that the filtering operation has already been completed after applying the filtering conditions. As a result, the model repeatedly clicks the filter button, leading to an execution loop and eventual task failure.

During our experiments, we implement an \textbf{early stop} mechanism: if the model outputs the same operation five consecutive times, the task is marked as failure in advance. We record the proportion of uncompleted tasks among all failed tasks, as well as the proportion of early-stopped tasks among all uncompleted tasks. The detailed results are presented in Table~\ref{table:uncompleted}. Across all models, the early stop ratio remains high. This phenomenon indicates that when the model repeatedly performs the same action, it lacks the ability to recognize and reflect on its ineffective behavior and make appropriate adjustments.

Additionally, we observe that GUI-specific models frequently encounter difficulties with \textit{"search"}, a crucial part in practice. The screenshots before and after focusing on the search box are often similar, making it challenging for the model to distinguish whether the focusing action has been executed, then repeatedly attempt to click the already focused search bar consequently. It further reflects the lack of attention and understanding of historical operation records. If the model could effectively review its historical behavior, it would be able to continue to enter the search content smoothly, thereby improving the overall task success rate.

\begin{table}[]
\centering
\renewcommand{\arraystretch}{1.2}
\setlength{\tabcolsep}{10pt}
\begin{tabular}{lcc}
\toprule
\textbf{Model}       & \textbf{Uncompleted}     & \textbf{Early Stop}    \\ \midrule
\multicolumn{3}{l}{\cellcolor[HTML]{EFEFEF}\textit{Proprietary Models}}   \\
Gemini 2.5 Pro       & 90.0                         & 49.6                       \\\midrule
\multicolumn{3}{l}{\cellcolor[HTML]{EFEFEF}\textit{General Open-source Models}}   \\
Qwen2.5-VL-7B       & 93.7                         & 63.7                       \\
Qwen2.5-VL-32B       & 92.3                         & 67.1                       \\
Qwen2.5-VL-72B       & 71.5                         & 50.0                       \\
InternVL3-8B       & 59.2                         & 77.9                       \\\midrule
\multicolumn{3}{l}{\cellcolor[HTML]{EFEFEF}\textit{GUI-specific Models}} \\
UI-TARS-1.5-7B       & 98.0                         & 43.8                       \\
GUI-R1-3B            & 25.6                         & 58.8                      \\
\bottomrule
\end{tabular}
\caption{The proportion of uncompleted tasks among failed tasks and early-stopped tasks among all uncompleted tasks. GPT-4o, Claude 4 Sonnet and UI-R1-E-3B are not included because of the poor accuracy.}
\label{table:uncompleted}
\end{table}

\subsection{Oversimplified Task Planning}
With the improvement of agents, users increasingly incline to issue more advanced and abstract task instructions, which is aligned with the instruction format in \model. In contrast, oversimplified task planning has become a significant bottleneck, limiting agents' potential for further breakthroughs. When confronted with complex and high-level requirements, agents should be capable of decomposing abstract instructions into a series of potential subtasks, leveraging the perception of the interface to dynamically adjust or execute specific subtasks, and efficiently driving the overall task towards completion.

However, we observe that most agents still demonstrate relatively simplistic task planning. In \model, agents frequently equate some query scenarios solely with interactions involving the search box. Specifically, when the current page does not allow direct task progression, they tend to input the entire instruction into the search box, trying to obtain the result in a single step. While this naive strategy may be effective for straightforward and direct tasks, it is generally inadequate for tackling more complex ones.

For example, consider the task \textit{"Go to the exchange rate conversion to see how many euros 100 Hong Kong dollars can be exchanged for"}, a natural and efficient task decomposition would involve first locating and opening the mini-program related to exchange rate conversion (either by searching or scrolling through the mini-program list), and then sequentially selecting the currency and entering the required amount until the desired information is obtained. 
However, the current planning capability cannot decompose such tasks in a reasonable, targeted, and hierarchical manner. Most models choose to directly enter \textit{"how many euros can 100 Hong Kong dollars be exchanged for"} into the search box, overlooking critical steps such as finding the correct program and gradually configuring conditions. This highlights notable shortcomings in multi-step task decomposition and execution strategy.

\section{Conclusion}
In this paper, we introduce \model, a comprehensive mobile benchmark with over 200 challenging GUI tasks covering widely-used scenarios. retaining the traditional State-related Task evaluation, we include Process-related Task and design a specialized method for accurate and efficient evaluation.
Our evaluation of advanced GUI agents indicates that even the best-performing models successfully complete less than 50\% of the tasks, and generally underperform on social and lifestyle applications. These findings underscore the need for advancements in grounding capability, historical operation attention, and task planning among existing agents. \model\ also has limitations. For instance, future work could improve the evaluation metrics to capture degrees of task progress rather than relying solely on binary judgments. Nevertheless, \model\ aspires to establish a new standard for evaluating operation tasks more accurately in mobile environments.

\section*{Acknowledgments}
This work was supported by the National Natural Science Foundation of China (Grant No.62372408).This work was supported by Ant Group Research Fund.

\bibliography{aaai2026}

\appendix

\section{\model\ Details}
\subsection{Implementation of Evaluator}
\begin{figure*}
    \centering
    \includegraphics[width=\linewidth]{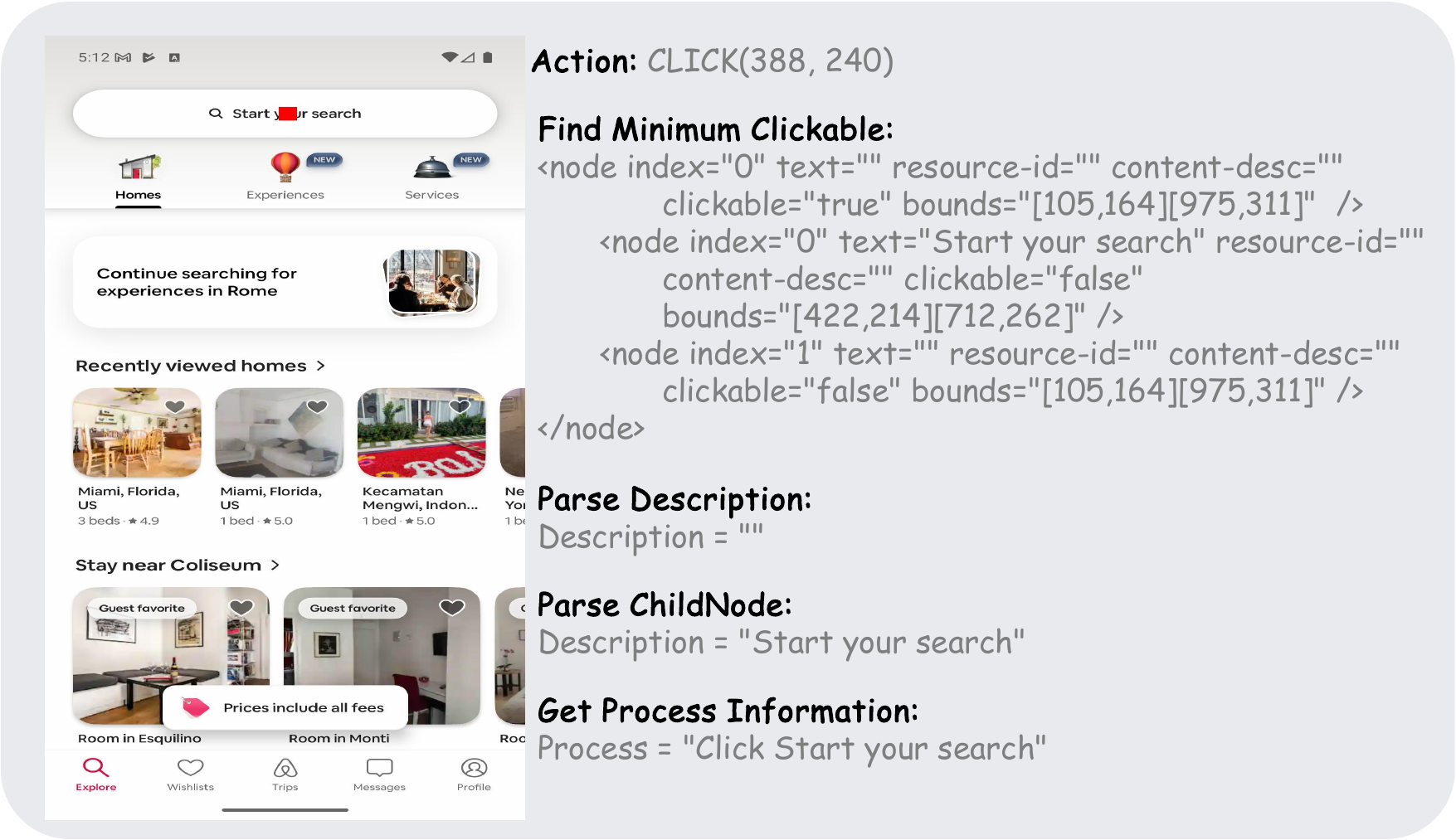}
    \caption{The execution process of Structure Description Converter to capture process information.}
    \label{fig:SDC}
\end{figure*}

Taking the example of clicking the search box on the Airbnb homepage, we show the execution process of \textbf{Structure Description Converter} in Figure~\ref{fig:SDC}. And the prompt for \textbf{MLLM-based Summarizer} is illustrated in the colored box below.
Meanwhile, we also show the prompts for evaluator to evaluate State-related Task and Process-related Task.

\section{Experimental Details}
For models like Gemini 2.5 Pro and InternVL3 which output coordinates ranging from 0 to 1000, we scale the coordinates according to the original image size for accurate click. For large-parameter models with strong command following capabilities, such as GPT-4o, Claude 4 Sonnet, Gemini 2.5 Pro and Qwen2.5-VL with 32B and 72B, we could obtain the expected instructions through simple action space hints. We choose GUI-R1-3B's prompt as the template while making some changes to apply to smaller agents which do not have official documentations. And we slightly modify all these prompts to fit our action space. All experimental GUI agents' prompts are shown as follows.

\section{Case Study}
In this part, we present more examples during the evaluation, in order to provide a deeper understanding of limitations in current agents.
\subsection{State-related Task}
\begin{figure*}
    \centering
    \includegraphics[width=\linewidth]{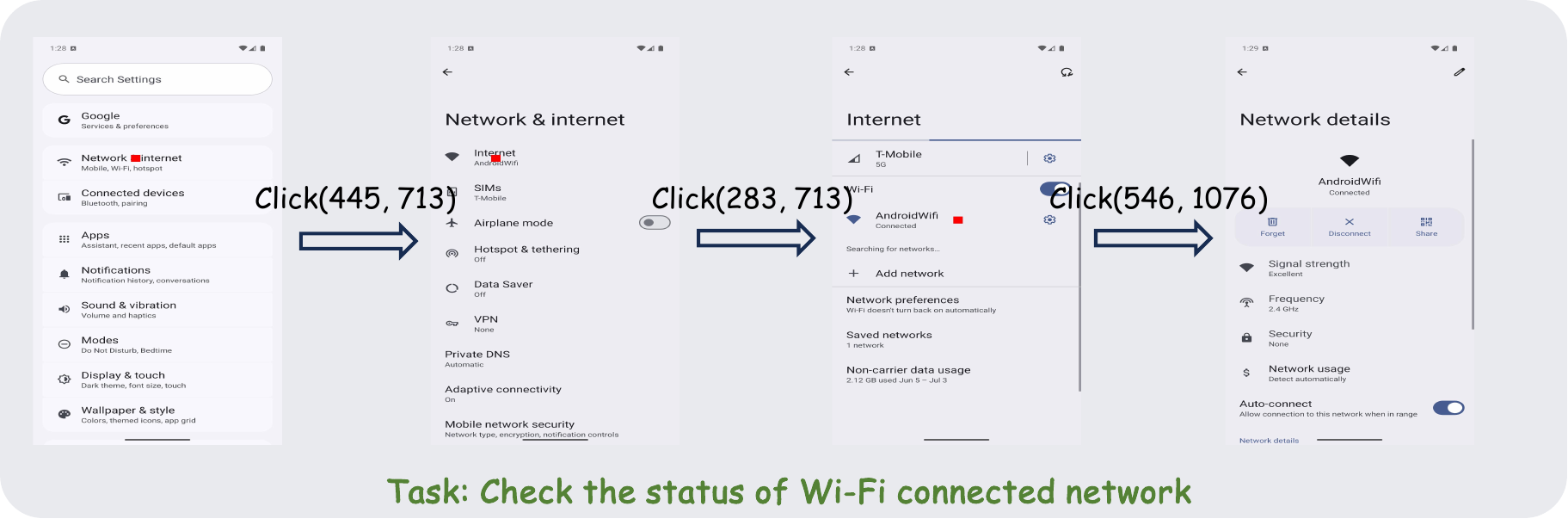}
    \caption{UI-TARS-1.5-7B's correct case in Settings of State-based Task.}
    \label{fig:state_true}
\end{figure*}

\begin{figure*}
    \centering
    \includegraphics[width=\linewidth]{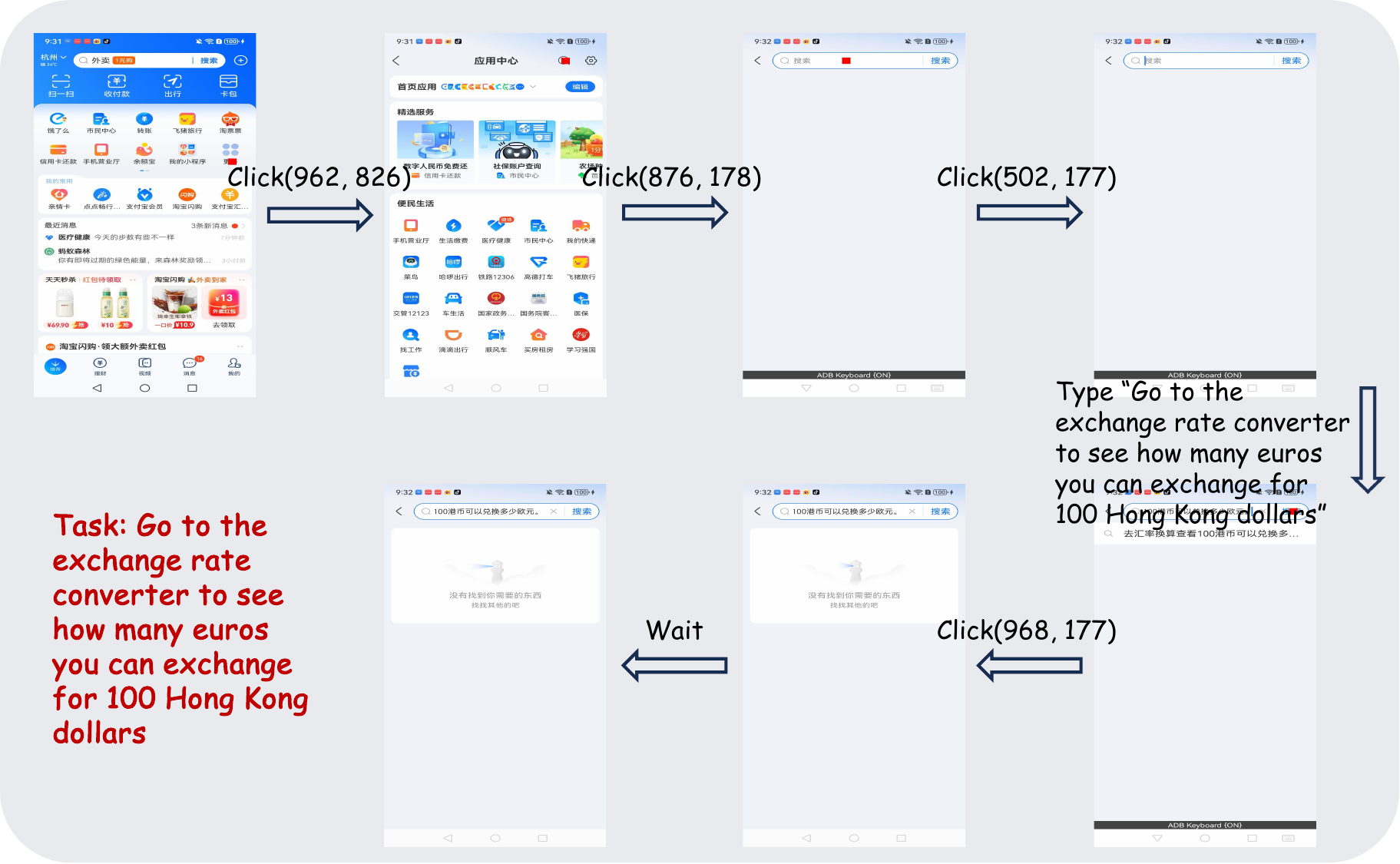}
    \caption{GUI-R1-3B's error case in Alipay of State-based Task.}
    \label{fig:state_false}
\end{figure*}

Figure~\ref{fig:state_true} shows a correct case of UI-TARS-1.5-7B to check the status of Wi-Fi connected network in Settings. We can figure out that, even small-parameter agents can successfully complete State-related Task consisting of multiple simple click actions. But in Figure~\ref{fig:state_false}, GUI-R1-3B is unable to complete more complex State-related Task. The model can only enter the entire question in the search box in the hope of obtaining the desired answer through the query. However, the real solution requires breaking down the complex instruction, which highlights the shortcomings of the existing task planning capability.
\subsection{Process-related Task}
\begin{figure*}
    \centering
    \includegraphics[width=\linewidth]{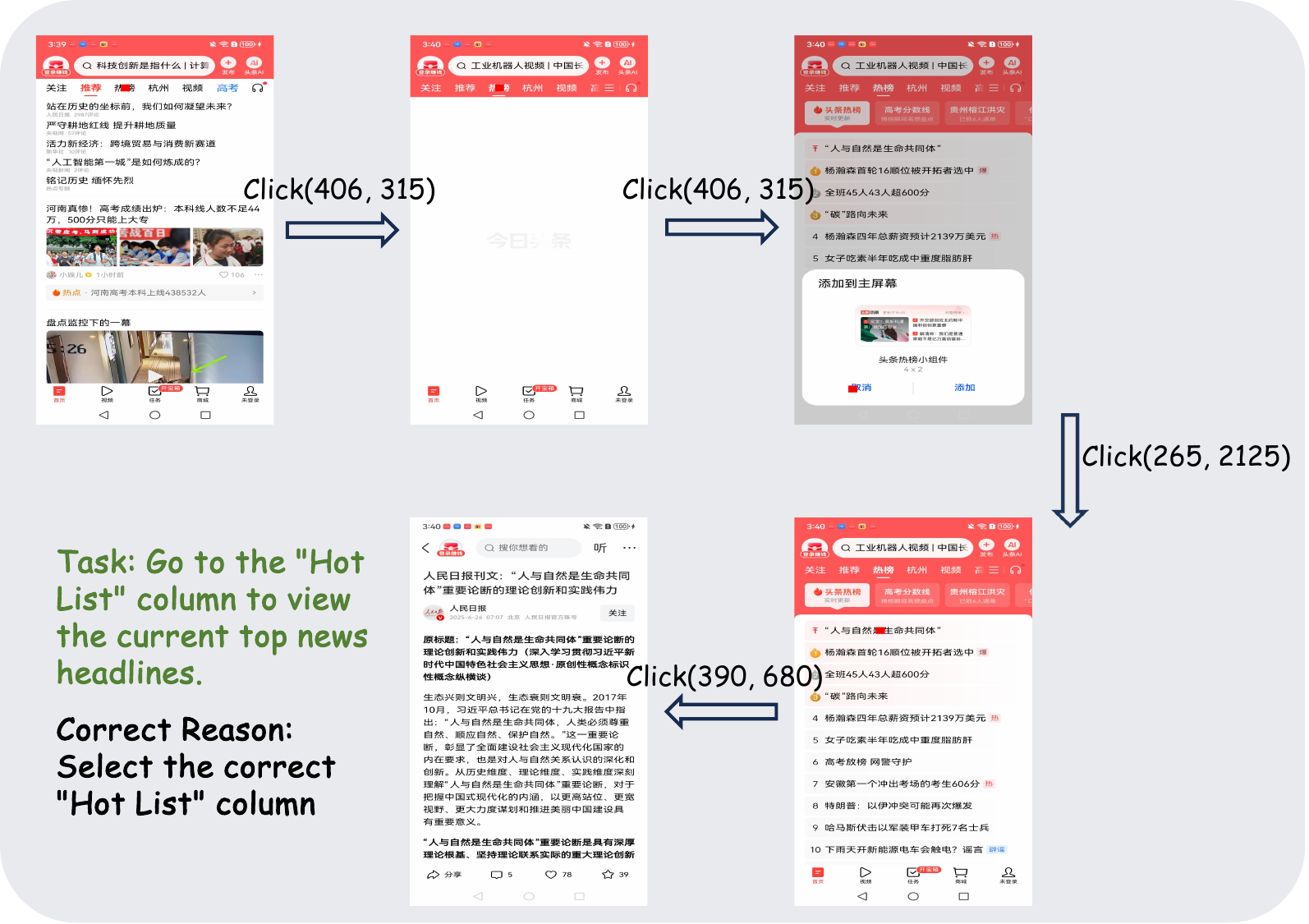}
    \caption{Qwen2.5-VL-72B's correct case in Toutiao of Process-based Task.}
    \label{fig:process_true}
\end{figure*}

The situation is even more challenging for more difficult Process-related Task that requires attention to process information. Since agents rarely succeed on longer tasks, we selected one of the few shorter successes. As shown in Figure~\ref{fig:process_true}, Qwen2.5-VL-72B successfully completed the task \textit{"Go to the Hot List column to view the current top news headlines"}, where selecting the Hot List column over other columns is crucial for success. However, this doesn't prove that agents can easily solve Process-related Task. In this task, the key operation is more like a natural step between opening the application and clicking on a news item, rather than a critical step that the model recognizes as necessary. Because selecting a column is always required before selecting a news item, choosing a random column rather than the "Hot List" column shown in the task instruction is almost impossible, leading to the final success. 

\begin{figure*}
    \centering
    \includegraphics[width=\linewidth]{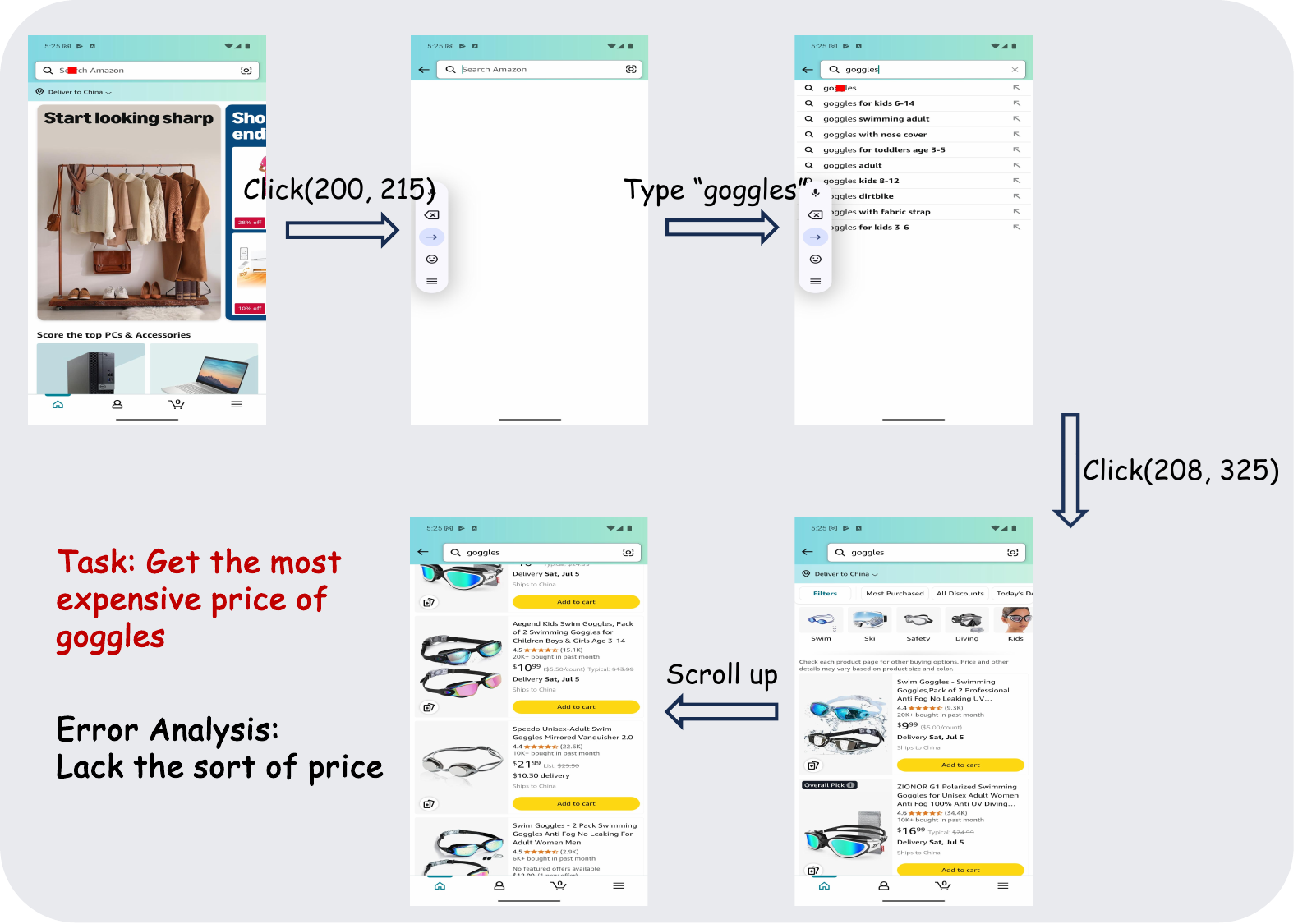}
    \caption{Gemini 2.5 Pro's error case in Amazon of Process-based Task.}
    \label{fig:process_false}
\end{figure*}

We also selected a task with similar length to highlight this issue. As illustrated in Figure~\ref{fig:process_false}, even the top-performing Gemini 2.5 Pro ignores the implicit price sorting requirement in the task instruction and simply scrolls through the page, believing it has found the cheapest goggles. Therefore, the Process-related Task introduced in our \model\ clearly demonstrates the lack of GUI agents' ability to rigorously capture and execute the necessary operation process.

\begin{table*}
\begin{tabular}{ll}
    \toprule
    \textbf{Application} & \textbf{Task Instruction}\\
    \midrule
airbnb & Get the search results near 'wembley stadium' for 1 adult.\\ 
airbnb & Get the list of available stays in Paris for 2 adults from October 10th to October 12th.\\ 
airbnb & Find experiences in Rome that include a guided tour of the Colosseum.\\ 
calculator & Get the result for factorial 7.\\ 
calculator & Get the result for sqrt(169).  \\ 
calculator & Calculate the value of cos(60°).  \\ 
calculator & Evaluate the expression (8+2)*(5-3).  \\ 
calculator & What is the result of $5^3$?  \\ 
calculator & Calculate 25\% of 300.  \\ 
chrome & Find the official website of Taobao.\\ 
chrome & Search for the current weather in Sydney, Australia.\\ 
chrome & Check the Chrome settings for Autofill options.\\ 
clock & Check what time it currently is in Tokyo, Japan.\\ 
clock & Check how many alarms are currently set in the app.\\ 
clock & Find out what the alarm sound is set to for the first alarm.\\ 
espn & Search for Klay Thompson. See all the articles and open one of them.\\ 
espn & Get the latest news about the New England Patriots.\\ 
espn & View the latest NBA match.\\ 
espn & Search for Lionel Messi and see his stats.\\ 
espn & See the current squad for Real Madrid.  \\ 
espn & Check the ESPN Power Rankings for college football.\\ 
gmail & Find out how many emails are currently in the Spam folder.\\ 
gmail & Check the status of Vacation Responder in the account settings.\\ 
gmail & Check current setting for the notification of Miscellaneous.\\ 
google\_maps & Search for nearby hotel rooms which can accommodate 4 adults and ratings higher than 4.\\ 
google\_maps & \makecell[l]{Get the search results for restaurants. Sort by distance and \\view the nearest American restaurant reviews.}\\ 
google\_maps & Check the days of the week when the Foster Museum is closed.\\ 
google\_play & Navigate to settings and check the status of notifications.\\ 
google\_play & Check the average user rating for Zoom\\ 
google\_play & Check the current version and last update date of Facebook.\\ 
settings & Check the current screen timeout.\\ 
settings & Check the current battery percentage and charging status.\\ 
settings & See which apps have been granted location permissions.\\ 
settings & Check the status of Wi-Fi connected network.\\ 
settings & View the current sound volume levels for media and calls.\\ 
settings & View the current storage space used by the Chrome.\\ 
yelp & \makecell[l]{Get the results for nearby restaurants. \\Filter to include only restaurants that offer takeout and sort by distance.}\\ 
yelp & Filter museums in New York City by those offering free WiFi.\\ 
youtube & Check all subscriptions and sort from A to Z\\ 
youtube & Retrieve the description of the channel @NationalGeographic\\ 
youtube & Get the list of posts in the Music category\\ 
youtube & Get the list of playlists created by the channel @TED-Ed\\ 
youtube & Retrieve the video details for How Earth Moves by @Vsauce\\ 
spotify & Find information about the album Midnights by Taylor Swift.\\ 
spotify & Look up details about the artist Billie Eilish.\\ 
spotify & Find the RapCaviar playlist and check its description.\\ 
spotify & Look up information about the podcast The Joe Rogan Experience.\\ 
spotify & Check the daily top 50 songs chart for the United States.\\ 
dictionary\_merriam\_webster & Check antonyms for generous.\\ 
dictionary\_merriam\_webster & What is the origin of the word algorithm?  \\ 
dictionary\_merriam\_webster & Check the first known use year of kangaroo.\\ 
dictionary\_merriam\_webster & Retrieve the examples for literally.\\ 
    \bottomrule
\end{tabular}
\caption{English State-related Task.}
\label{table:esq}
\end{table*}

\begin{table*}
\centering
\begin{tabular}{ll}
    \toprule
    \textbf{Application} & \textbf{Task Instruction}\\
    \midrule
    airbnb & Find apartments in New York City and filter with 3 bedrooms.\\ 
airbnb & Check the availability of a hotel in Bali with a pool and allow pets.\\ 
airbnb & \makecell[l]{View the distance from any listing near Central Park Apartment \\to the Empire State Building.}\\ 
airbnb & Retrieve the amenities included in the first search result in Miami.\\ 
airbnb & View the reviews from experiencs of Harry Potter's London.\\ 
amazon & Get the most expensive price of goggles\\ 
amazon & Get the product description for AmazonBasics AA Rechargeable Batteries\\ 
amazon & Get the reviews information for TP-Link WiFi Router.\\ 
chrome & Find a news article about artificial intelligence.\\ 
espn & Look up the top 10 fantasy baseball prospects.  \\ 
google\_maps & \makecell[l]{Search for a nearby gas station that is open now. \\Set it as the final destination while McDonald as the first stop.}\\ 
google\_maps & Search for a 24-hour access gym and check the walking time to the location.\\ 
google\_maps & Search for coffee shops good for kids and ratings higher than 4.5.\\ 
google\_maps & View the bus route from the current location to the nearest bank.\\ 
google\_play & Search for WhatsApp and sort the reviews by most recent.\\ 
google\_play & Find apps similar to Instagram.\\ 
google\_play & Browse the game screenshots of Jelly Crush Saga app.\\ 
yelp & Find the highest rated sushi restaurant in Tokyo and check its full menu.\\ 
yelp & Get newest reviews for the Eiffel Tower from travelers.\\ 
yelp & Get the opening hours of the haircut at top-rated barbershops in London.\\ 
youtube & Search videos about LeBron James. Check the comments of one of the results\\ 
youtube & Get the list of videos of Comedy sorted by most views\\ 
dictionary\_merriam\_webster & Check the definition of one synonym of agent.\\ 
    \bottomrule
\end{tabular}
\caption{English Process-related Task.}
\label{table:epq}
\end{table*}

\onecolumn
\begin{tcolorbox}[colback=brown!7, title = {Prompt for MLLM-based Summarizer}]
{\tt
You will receive a picture that is a horizontal stitching of two pictures.\\

The left side of the red dividing line represents the screen before the click operation, and the right side represents the screen after it.

Besides, I will give you the original coordinate of the click.\\

You need to analyze what this operation has actually done on the screen, based on given information and 
    the changes in the screens (for example, open a certain app, click a certain button).\\

Here is the original description of the operation: <action>\\

You are required to summarize this operation with a verb phrase that begins with the given operation type. 

If the operation does not cause any changes to the two images, output Invalid click.\\

Show your thinking process wrapped in <think> </think>. And output the summary wrapped in <summary> </summary>.}
\end{tcolorbox}

\begin{tcolorbox}[colback=black!7, title = {Prompt for Evaluation of State-related Task}]
{\tt             
You are an expert in smartphone task evaluation.

I will give you a query task. Your responsibility is to determine whether the current image could answer the query task.

You need to carefully compare the task goal with the information on the image. Only if information could answer the task completely, the judgment success.\\

The task is:
<goal>

Show your thinking process wrapped in <think> </think>.
Output True or False wrapped in <answer> </answer>}
\end{tcolorbox}

\begin{tcolorbox}[colback=black!7, title = {Prompt for Evaluation of Process-related Task}]
{\tt             
You are an expert in smartphone task evaluation.

I will give you a query task and some execution information during the operation. Your responsibility is to determine whether the current image could answer the query task.

You need to carefully compare the task goal with the information on the image. At the same time, you need to pay special attention to whether the process information can meet the task requirements that cannot be shown in the image, such as sorting by distance, completing filtering, etc.

The task is successful only if all its requirements are met.\\

The task is:
<goal>

The process information is:
<process>

Show your thinking process wrapped in <think> </think>.
Output True or False wrapped in <answer> </answer>}
\end{tcolorbox}

\begin{tcolorbox}[colback=blue!7, title = {Prompt for GPT-4o, Claude 4 Sonnet, Gemini 2.5 Pro and Qwen2.5-VL-72B}]
{\tt You will receive the current screen image.

Your overall goal is: <goal> And you need to complete it within current app.

Historical actions you have performed: <history>\\

These are the action space to interact with the phone:

- Click(x, y): Click a coordinate point on the screen and x, y is the position of the coordinate point.

Click(100,238) means click the UI element at (100,238) on the current screen.

- Type(text): Type text.

- Swipe(x1, y1, x2, y2): Scroll the screen from point A to point B. The coordinates of point A are x1, y1, and the coordinates of point B are x2, y2.

Swipe(300,800,300,200) swipes the screen from (300, 800) to (300, 200).

- Back(): Return to the previous step.

- Enter(): Pressing the ENTER key to submit.

- Wait(): Wait a while for the network to load.

- Complete(): It means you think the task is completed.\\

Now according to the above guidance and the screen state, think step-by-step about the action that should be done.\\

You can only answer actions in the "action space". Output only one action at a time and follow the format in the "action space".
Start with "Action:" and do not output any other thought process!}
\end{tcolorbox}

\begin{tcolorbox}[colback=blue!7, title = {Prompt for Qwen2.5-VL-32B}]
{\tt             
You will receive the current screen image.

Your overall goal is: <goal> And you need to complete it within current app.

Historical actions you have performed: <history>\\

These are the action space to interact with the phone:

- Click(x, y): Click a coordinate point on the screen and x, y is the position of the coordinate point.

Click(100,238) means click the UI element at (100,238) on the current screen.

- Type(text): Type text.

- Swipe(x1, y1, x2, y2): Scroll the screen from point A to point B. The coordinates of point A are x1, y1, and the coordinates of point B are x2, y2.

Swipe(300,800,300,200) swipes the screen from (300, 800) to (300, 200).

- Back(): Return to the previous step.

- Enter(): Pressing the ENTER key to submit.

- Wait(): Wait a while for the network to load.

- Complete(): It means you think the task is completed.\\

Now according to the above guidance and the screen state, think step-by-step about the action that should be done.\\

Output the thinking process in <think> </think> tags, and the final answer in <answer> </answer> tags as follows:\\
<think> ... </think> <answer>Swipe(x1, y1, x2, y2)</answer>

You must follow the format in "action space", for example:
    Click(1000, 2000)
    Swipe(1000, 500, 1000, 1480)
    Type(text)
    Back()
    Enter()
    Wait()
    Complete()\\
Output only one action at a time.}
\end{tcolorbox}

\begin{tcolorbox}[colback=blue!7, title = {Prompt for Qwen2.5-VL-7B, InternVL3-8B, GUI-R1-3B}]
{\tt In this UI screenshot, I want you to continue executing the command <goal>, with the action history being <history>.\\

Please provide the action to perform (enumerate from ['wait', 'complete', 'click', 'back', 'type', 'enter', 'scroll']), the point where the cursor is moved to (integer) if a click is performed, and any input text required to complete the action.\\

Output the thinking process in <think> </think> tags, and the final answer in <answer> </answer> tags as follows:

<think> ... </think> <answer>[\{'action': enum['wait', 'complete', 'click', 'back', 'type', 'enter', 'scroll'], 'point': [x, y], 'input\_text': 'no input text [default]'\}]</answer>\\

Note: 

specific input text (no default) is necessary for actions enum['type', 'scroll'] and only output one action at a time\\

Example:

    [\{'action': enum['wait', 'back', 'complete', 'enter'], 'point': [-100, -100], 'input\_text': 'no input text'\}]
    
    [\{'action': enum['click'], 'point': [123, 300], 'input\_text': 'no input text'\}]
    
    [\{'action': enum['type'], 'point': [-100, -100], 'input\_text': 'shanghai shopping mall'\}]
    
    [\{'action': enum['scroll'], 'point': [-100, -100], 'input\_text': enum['up', 'left', 'right', 'down']\}]}
\end{tcolorbox}

\begin{tcolorbox}[colback=blue!7, title = {Prompt for UI-TARS-1.5-7B}]
{\tt
You are a GUI agent. You are given a task and your action history, with screenshots. You need to perform the next action to complete the task. Only output one action at a time.\\

\#\# Output Format

Thought: ...

Action: ...\\

\#\# Action Space

click(point='<point>x1 y1</point>')

type(content='')

drag(start\_point='<point>x1 y1</point>', end\_point='<point>x2 y2</point>')

press\_back()

press\_enter()

finished()\\

\#\# Note

- Write a small plan and finally summarize your next action (with its target element) in one sentence in `Thought` part.\\

\#\# User Instruction

<goal>\\

\#\# Action History

<history>
}
\end{tcolorbox}

\begin{tcolorbox}[colback=blue!7, title = {Prompt for UI-R1-E-3B}]
{\tt In this UI screenshot, I want to perform the command <goal>.\\

Please provide the action to perform (enumerate in ['wait', 'complete', 'click', 'back', 'type', 'enter', 'scroll']), the point where the cursor is moved to(integer) if click is performed, and any input text required to complete the action.\\

Output the thinking process in <think> </think> and final answer in <answer> </answer> tags.\\

The output answer format should be as follows:

    <think>...</think> <answer>[{'action': enum['wait', 'complete', 'click', 'back', 'type', 'enter', 'scroll'], 'point': [x, y], 'input\_text': 'no input text [default]'}]</answer>
    
Please strictly follow the format.}
\end{tcolorbox}

\twocolumn

\end{document}